\documentclass{article}
\usepackage{ijcai21}

\usepackage{times}
\usepackage{soul}
\usepackage{url}
\usepackage[hidelinks]{hyperref}
\usepackage[utf8]{inputenc}
\usepackage[small]{caption}
\usepackage{graphicx}
\usepackage{amsmath}
\usepackage{amsthm}
\usepackage{booktabs}
\usepackage{algorithm}
\usepackage{algorithmic}
\usepackage{color}
\usepackage{enumitem}
\usepackage{subfigure}
\usepackage{amsfonts}       

\title{Hidden State Approximation in Recurrent Neural Networks Using Continuous Particle Filtering}

\author{
Dexun Li
\affiliations
Singapore Management University\\
\emails
dexunli.2019@phdcs.smu.edu.sg
}

\begin{document}

\maketitle

\begin{abstract}
Using historical data to predict future events has many applications in the real world, such as stock price prediction; the robot localization. In the past decades, the Convolutional long short-term memory (LSTM) networks have achieved extraordinary success with sequential data in the related field. However, traditional recurrent neural networks (RNNs) keep the hidden states in a deterministic way. In this paper, we use the particles to approximate the distribution of the latent state and show how it can extend into a more complex form, i.e., the Encoder-Decoder mechanism. With the proposed continuous differentiable scheme, our model is capable of adaptively extracting valuable information and updating the latent state according to the Bayes rule. Our empirical studies demonstrate the effectiveness of our method in the prediction tasks.
\end{abstract}

\section{Introduction}
Using sequential data to predict future value has been a popular topic for decades, especially in the financial market, i.e., stock price prediction. During the last several decades, numerous time-varying models have been developed and applied extensively to characterize the dynamic evolution of the hidden state that is inherent in time series data and use it to predict future value. However, most of the non-linear models, such as the Stochastic Volatility Model~\cite{hull1987pricing,taylor1994modeling,malik2011particle}, where the likelihood function cannot be evaluated analytically and may be numerically formidable. Consequently, traditional methods such as the standard maximum likelihood estimation method cannot be applied directly, and the predefined nonlinear form may lose the flexibility to capture the various nonlinear relationships. Meanwhile, the traditional forecasting methods make use of both linear (AR, MA, ARIMA) and non-linear algorithms (ARCH, GARCH, SVM), but they only focus on the target series, and ignore the relevant driving series. To address the aforementioned problem, state-of-the-art (SOTA) sequence predictors take the advantage of the RNNs to keep track of the belief of the hidden state while can adaptively capture the underlying nonlinear relationship among the exogenous input terms. RNNs have shown their success in time series prediction. However, RNNs use a deterministic representation of the hidden state, which accumulates the uncertainty as the time horizon increases and lead to poor learning results. Instead, we use a set of particles to represent the hidden state variable, and update it in a Bayesian manner. This belief representation enables our model to better handle the time series prediction with uncertainty. 


The major contributions of this paper are summarized as follows:
\begin{itemize}
    \item We develop a novel Continuous Particle Filtering LSTM (CPF-LSTM) model which enables continuous likelihood approximation which can be negated in the optimization process, and we provide an ELBO in the research domain.
    \item We show how our proposed method can be incorporated into the more complex extension models straightforwardly.
    \item We compare our model with SOTA to evaluate the effectiveness of our method on the NASDAQ 100 Stock dataset with a set of driving series. Our model achieves an improvement compared with SOTA in terms of mean square error (MSE).
\end{itemize}

\section{Background}
\subsection{Sequence Data Prediction}
The autoregressive conditional heteroscedasticity (ARCH) model was introduced in Engle~\shortcite{engle1982autoregressive} to characterize the dynamic evolution of the volatility that is inherent in financial time series data and is extended to generalized ARCH (GARCH) by Bollerslev~\shortcite{bollerslev1986generalized}. As an alternative to the ARCH framework, the stochastic volatility (SV) model, originated by Taylor, is specified to follow some latent stochastic process and has been used as an approximation to the stochastic volatility diffusion~\cite{hull1987pricing,chesney1989pricing}. SV models have gradually emerged as a useful way to model a changing time series, especially in financial economics. 
However, those models only consider the target series value $(y_1, \dots, y_{t-1})$ and ignore the (exogenous) driving series value $(\mathbf{x}_1,\dots,\mathbf{x}_t)$. Instead, various attempts using the dynamical neural architecture have been made as the non-linear autoregressive exogenous models ~\cite{gao2005narmax,menezes2008long,ardalani2010chaotic,boussaada2018nonlinear}. Those input-output modeling of nonlinear dynamical systems show promising quality, for example, the LSTM model achieved great success in various applications. The output of the LSTM model depends on the output of the hidden layer of the previous time step along with the current input. Formally, given the input time series $\mathbf{X} =(\mathbf{x}^1,\mathbf{x}^2,\dots,\mathbf{x}^n)^\top = (\mathbf{x}_1,\mathbf{x}_2,\dots,\mathbf{x}_T)\in \mathbb{R}^{n\times T}$, where $T$ is the length of window size, and we use $\mathbf{x}^k = (x_1^k, x_2^k, \dots, x_T^k)$ to represent a driving series of length $T$ and employ $\mathbf{x}_t = (x_t^1, x_t^2, \dots, x_t^n)^\top$ to denote a vector of $n$ exogenous (driving) input series at time $t$. LSTM updates as follows:


\begin{align}
    \mathbf{f}_t &= \sigma (\mathbf{W}_f \cdot [\mathbf{h}_{t-1}; \mathbf{x}_t] + \mathbf{b}_f) \label{eq:lstm_1}\\
    \mathbf{i}_t &= \sigma (\mathbf{W}_i \cdot [\mathbf{h}_{t-1}; \mathbf{x}_t]+ \mathbf{b}_i) \label{eq:lstm_2}\\
    \mathbf{o}_t &= \sigma (\mathbf{W}_o \cdot [\mathbf{h}_{t-1}; \mathbf{x}_t] + \mathbf{b}_o) \label{eq:lstm_3}\\
    \tilde{\mathbf{C}}_t &= \tanh (\mathbf{W}_C \cdot [\mathbf{h}_{t-1}; \mathbf{x}_t]+ \mathbf{b}_C) \label{eq:lstm_cell}\\
    \mathbf{C}_t &= \mathbf{f}_t \ast \mathbf{C}_{t-1} +\mathbf{i}_t \ast \tilde{\mathbf{C}}_t \label{eq:lstm_5}\\
    \mathbf{h}_t &= \mathbf{o}_t \ast \tanh (\mathbf{C}_t) \label{eq:lstm_trans}
\end{align}
where $\mathbf{h}_{t-1}\in\mathbb{R}^{m}$ is the hidden state at time $t-1$, $m$ is the dimension of the hidden state, and $[\mathbf{h}_{t-1}; \mathbf{x}_t]$ is the concatenation of the previous hidden state $\mathbf{h}_{t-1}$ and the current input $\mathbf{x}_t$. $\mathbf{W}_f, \mathbf{W}_i, \mathbf{W}_o, \mathbf{W}_C\in \mathbb{R}^{m\times (m+n)}$ are the weight matrices and $\mathbf{b}_f,\mathbf{b}_i,\mathbf{b}_o,\mathbf{b}_C\in\mathbb{R}^m$ are bias vectors. $\sigma$ and $\ast$ represent logistic sigmoid function and element-wise multiplication, respectively.

\subsection{Particle Filtering}

Particle filtering (PF), also known as sequential Monte Carlo methods, is a technique by using a recursive filter by Monte Carlo simulations, extending the Kalman filter to non-linear and non-Gaussian state-space models. The recursive propagation can be approximated by the corresponding empirical density. As the particle filtering in coupled with bootstrap only requires having to simulate forward in time from the transition density of the unobserved states, it is typically straightforward. The general approach recursively delivers sequences of samples $\{ h_t\}$ from the distributions under parameters $\theta$:
\begin{equation*}
 p(h_1|Y_{0},\theta),\dots,p(h_t|Y_{t-1},\theta),
\end{equation*}
where $Y_t$ is contemporaneously available information. A non-Gaussian state-space approach to the modeling of non-stationary time series is provided by Kitagawa~\shortcite{kitagawa1987non}. Gordon~\shortcite{gordon1993novel} and Kitagawa~\shortcite{kitagawa1996monte} developed the general framework of sampling importance resampling (SIR) algorithm. The approach is to represent the required density by a set of random particles with associated weights, and Bayesian theory is used repeatedly to re-weight the particles in advancing the system. Such a filter consists of essentially two stages: prediction and updating. From Bayes theorem,
\begin{equation}
p(h_t|Y_t,\theta)\propto p(y_t|h_t,\theta)p(h_t|Y_{t-1},\theta),
\end{equation}
where
\begin{equation}
p(h_t|Y_{t-1},\theta)=\int p(h_t|h_{t-1},\theta)p(h_{t-1}|Y_{t-1},\theta)dh_{t-1},
\end{equation}
and this integral can be approximated by a recursive filtering approach using $K$ particles $\{h_t^1,\dots,h_t^K\}$. It means that received data can be processed sequentially rather than as a batch so that it is not necessary to store the complete data, and not to reprocess existing data if a new measurement becomes available. The objective of the prediction is to obtain $p(h_t|Y_t)$ which can be approximated by the empirical prediction density. It can be given according to the Chapman-Kolmogorov equation:
\begin{equation}\label{Prediction}
p(h_t|Y_{t-1},\theta) \simeq \frac{1}{K} \sum_{k=1}^K p(h_t|h_{t-1}^k,\theta)
\end{equation}
The updating stage which can be obtained via the Bayes rule is approximated by the empirical filtering density:
\begin{equation}
\begin{aligned}
    f(h_{t+1}|Y_{t+1}) &\propto f(y_{t+1}|h_{t+1})\int f(h_{t+1}|h_t)f(h_t|Y_t)d h_t \\
    &\simeq p(y_t|h_t,\theta) \frac{1}{K} \sum_{k=1}^K p(h_t|h_{t-1}^k,\theta).
\end{aligned}
\end{equation}

We now summarize the basic sample and resampling steps of PF algorithm in period $t$ as follows:
\begin{enumerate}
\item  \textbf{Transition update process}\label{PF1}. Given $K$ particles $\{h_{t-1}^1,\dots,h_{t-1}^K\}$ from $p(h_{t-1}|Y_{t-1};\theta)$, evolve particles according to the transition probability density
\[
\widetilde{h}_{t}^k \sim p(h_t|h_{t-1}^k,\theta)
\]
\item \textbf{Measurement update process}\label{PF2}. For $k=1:K$, calculate the normalised weights
\[
\pi_{t}^k=\frac{\omega_{t}^k}{\sum_{i=1}^{K}\omega_{t}^i}, \quad where\quad   \omega_{t}^k=p(y_t|\widetilde{h}_{t}^k) 
\]
\item \textbf{Resampling process}\label{PF3}. For $k=1:N$, resample among $\{\widetilde{h}_{t}^1,\dots,\widetilde{h}_{t}^K\}$ with probability proportional to $\{\pi_{t}^1,\dots \pi_{t}^K\}$ to update filtered sample $\{h_{t}^1,\dots,h_{t}^K\}$ from $p(h_t|Y_t;\theta)$.
\[
h_{t}^k \sim \sum_{k=1}^K \pi_t^k \delta(h_t-\widetilde{h}_{t}^k)
\]
\end{enumerate}
Here, we use $\delta(\cdot)$ to denote the Dirac-delta measure. As $t$ varies, this will yield an approximation of the desired posterior density of $p(h_t|Y_t)$.



\section{RNNs with Particle Filtering}\label{sec:PF_RNN}
In this section, we will demonstrate how to use a continuous particle filtering algorithm in conjunction with an LSTM model to approximate the hidden state (belief) distribution with a set of weighted latent particles. More specifically, we will demonstrate how to negate our proposed model to SOTA.


\subsection{Continuous Particle Filtering for RNNs}
We want to maintain a batch of particles to approximate the distribution of the latent state, while the traditional latent state is kept in a deterministic way. There are two steps in the particle filtering algorithm: transition update and measurement update. In the LSTM unit, we will demonstrate how to address these two steps in detail.
\subsubsection{Transition Update}
we apply the transition function (Eq.~\ref{eq:lstm_trans}) to each particle hidden state $\{ h_t^i \}$. Similar to the existing work~\cite{ma2020particle}, we add the noise term into the equation to increase the particle diversity and relieve the issue of particle depletion after resampling.
\begin{equation}\label{eq:pf_lstm}
    \tilde{\mathbf{h}}_t^i = \mathbf{o}_t^i\ast \tanh (\mathbf{C}_t^i) + \mathbf{\epsilon}_t^i
\end{equation}
We assume that $\mathbf{\epsilon}_t^i\in \mathbb{R}^m$ to be a learned Gaussian distribution, where $\mathbf{\epsilon}_t^i \sim f_{noise}(h_{t-1}^i, \mathbf{x}_t)$. 

\subsubsection{Measurement Update}
In the basic sampling importance re-sampling filter, we also need to calculate the measurement density $p(y_t|\tilde{\mathbf{h}}_t^i)$, where $y_t$ is the true predicted value. Instead of using a generative model to model the measurement density, we approximate its log value, $\log p(y_t|\tilde{\mathbf{h}}_t^i)$, by a learned function $f_{weight}(y_t,\mathbf{h}_t)$.

\subsubsection{Resampling Process}\label{subsub:resamp}
The original multinomial sampling scheme (also known as the weighted bootstrap) is computational $O(N\times K)$, where $N$ is the number of training samples utilized per iteration in the LSTM model. 

\subsubsection{Prediction}
We apply the prediction process by a learned function $\hat{y} = f_{out}(\bar{\mathbf{h}}_t)$, where the $\bar{\mathbf{h}}_t$ is the mean of the particle hidden states $\frac{1}{K}\sum_{i=1}^K \tilde{\mathbf{h}}_t^i$. 

\subsubsection{Training Procedure}
The standard way to train the model is to optimize the mean squared error, i.e., the form used by Qin {\it et al.}~\shortcite{qin2017dual}:
\begin{equation}
    \mathcal{O}(y_T,\hat{y}_T) = \frac{1}{N} \sum_{i=1}^N (\hat{y}_T^i - y_T^i)^2,
\end{equation}
where $N$ is the number of training samples. Motivated by ~\cite{burda2015importance,ma2020particle}, we propose another objective function for optimization. In particular, we use the following lower bound, corresponding to the weights of $K$ particles which is used to approximate the likelihood:

\begin{equation}
\begin{aligned}
    \log \hat{L}(\theta) &= \sum_{t=1}^T \log(\frac{1}{K} \sum_{i=1}^K  p(y_t^i|\theta))\\
    &= \sum_{t=1}^T \log ( \frac{1}{N} \sum_{i=1}^N(\frac{1}{K} \sum_{k=1}^K w_{t}^{i,k}) )
\end{aligned}
\end{equation}
where $w_{t}^{i,k}$ is the unnormalized importance weights for particles $k$ in the training samples $i$ at time $t$. This is a lower bound on the marginal log-likelihood, as follows from Jensen's Inequality and the fact that the average importance weights are an unbiased estimator of $p(h_t|Y_t)$:
\begin{equation}
    \begin{aligned}
    \mathcal{L}_{ELBO} &= \mathbb{E}\lbrack \sum_{t=1}^T \log ( \frac{1}{N} \sum_{i=1}^N(\frac{1}{K} \sum_{k=1}^K w_{t}^{i,k}) )  \rbrack \\
    &=\sum_{t=1}^T  \mathbb{E}\lbrack \log ( \frac{1}{N} \sum_{i=1}^N(\frac{1}{K} \sum_{k=1}^K w_{t}^{i,k}) ) \rbrack \\
    &\leq \sum_{t=1}^T \log \mathbb{E} \lbrack ( \frac{1}{N} \sum_{i=1}^N(\frac{1}{K} \sum_{k=1}^K w_{t}^{i,k}) ) \rbrack = \log \hat{L}(\theta)
    \end{aligned}
\end{equation}
More specifically, we can train over the combination of the two learning objectives as follows:
\begin{equation}
    \mathcal{O}(y_T,\hat{y}_T) + \kappa  \mathcal{L}_{ELBO}
\end{equation}
where $\kappa$ is the trade-off coefficient, measuring the relative weights of likelihood estimation and prediction accuracy in the training process.


\subsubsection{Continuous Resampling}
\begin{figure}[t]
  \centering
      \includegraphics[width=0.49\textwidth, height=1.5in]{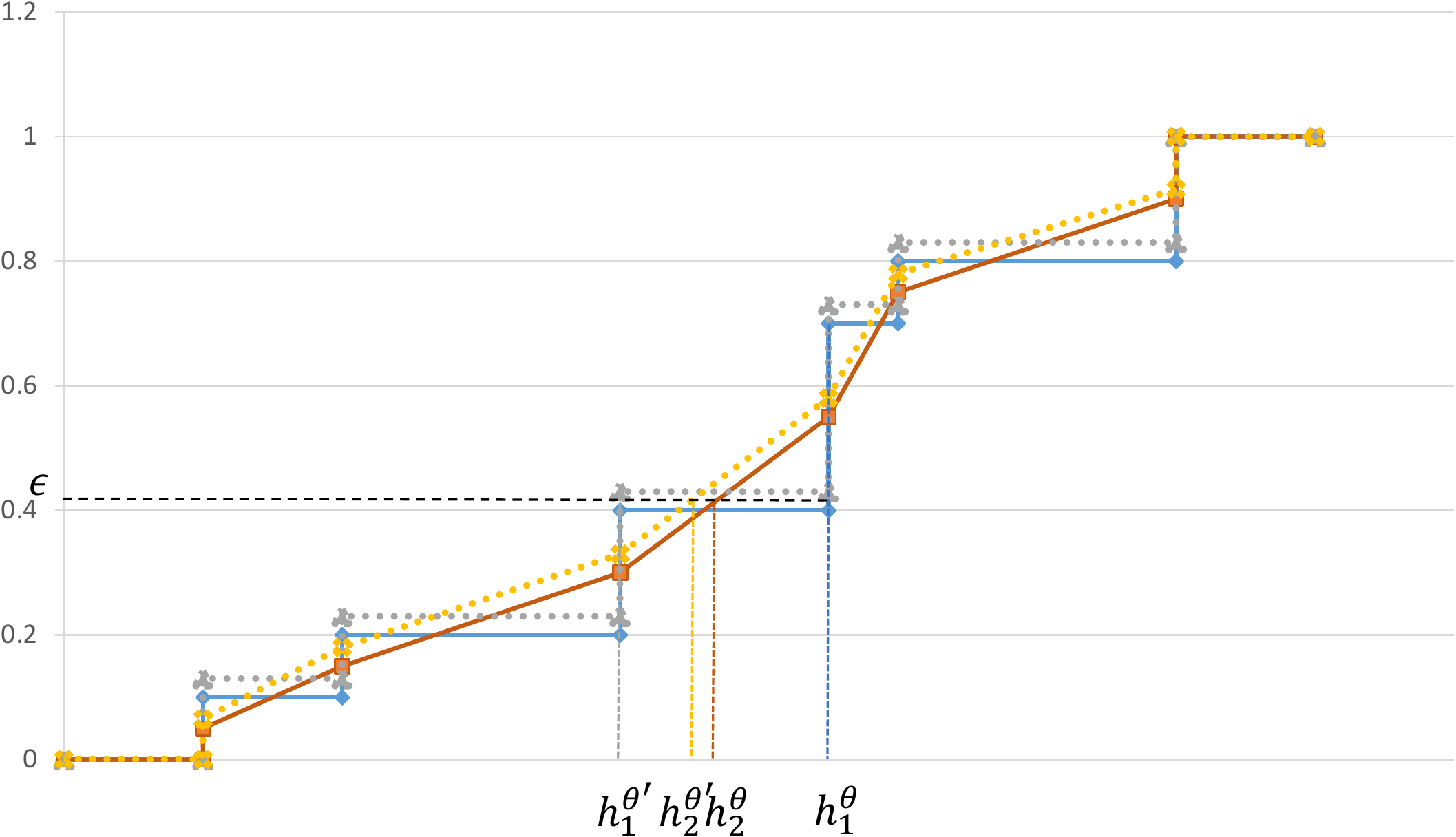}\\
  \caption{The blue solid line step function is the discontinuous empirical cumulative distribution function (ECDF) under some parameters $\theta$, and the red solid line is the continuous approximation of the ECDF under the same parameter $\theta$. The dashed grey line is the discontinuous ECDF under parameter $\theta^{\prime}$, where $\theta^{\prime}$ is very close to $\theta$, again, the dashed yellow line is the continuous approximation of the ECDF under $\theta^{\prime}$. By inverting these continuous approximations according to a fixed uniform $\epsilon$, large differences between sampled particles under a small change in the discontinuous resampling process are eliminated (from $||h_1^{\theta}-h_1^{\theta^{\prime}}||$ to $||h_2^{\theta}-h_2^{\theta^{\prime}}||$). }\label{fig:cont}
\end{figure}

However, even with the randomness fixed, evaluating the likelihood at different values of $\theta$ will not result in the construction of a smooth likelihood surface or the mean squared error. Discontinuities arise from the bootstrap \textit{resampling process} (see section~\ref{subsub:resamp}). Specifically, sampling the particles $h_t^k$ at the step (\ref{PF3}) in the PF algorithm from the weighted empirical cumulative distribution function (ECDF) leads to the discontinuities
\begin{equation}
\widehat{F}_K(h_t)=\sum_{k=1}^K\pi_t^k \mathbb{I} (h_t-\widetilde{h}_t) 
\end{equation}
where $\mathbb{I}(\cdot)$ is the indicator function. In order to solve this, Malik and Pitt~\shortcite{malik2011particle} propose constructing a continuous approximation $\widetilde{F}_K(h)$ of $\widehat{F}_K(h)$ and then resampling particles $\{ \widetilde{h}_t^k \}$ by inverting uniforms based on $\widetilde{F}_K(h)$. Inspired by their work, we propose a novel continuous resampling process that could be applied in the high dimensional domain. We first project the particles $\widetilde{h}_t^1,\dots,\widetilde{h}_t^K$ into the one dimension space. Intuitively, we use the learned $f_{out}()$ to get the projection of the hidden states, and we sort the particles according to the ascending order of the projection value, then we construct the continuous EDCF as follows:
\begin{equation}
    \widetilde{F}_K(h)=\lambda^0 \mathbb{I}(h\ge h_t^1)+\sum_{k=1}^{K-1}\lambda^k G_k \left( \frac{h-h_t^k}{h_t^{k+1}-h_t^k} \right)+\lambda^K\mathbb{I}(h\ge h_t^K),
\end{equation}
where $\lambda^0=\frac{\pi^{1}}{2}$, $\lambda^K=\frac{\pi^{K}}{2}$, $\lambda^k=\frac{\pi^{k}+\pi^{k+1}}{2}$ for $k=1,\dots,K-1$ and $\pi^k$ denotes the normalised weights $\pi_t^k$ in PF algorithm. The function $G_k(z)$ is chosen as a distribution function of $\lbrack0,1\rbrack$, specifically, $G_k(z)=0$ for $z\leq 0$ and $G_k(z)=1$ for $z\geq 1$ and $G_k(z)=z$ otherwise. Similar to \cite{malik2011particle}, the distance
\[
\lVert \widehat{F}_K(h)-\widetilde{F}_K(h) \rVert_\infty \ \leq \mathop{\sup}_{h} \lvert \widehat{F}(h)-\widetilde{F}(h) \rvert = \mathop{\max}_{k\in \{1,\dots,K\}}\lbrace \frac {\pi^k}{2} \rbrace,
\]
which is of order $\frac{1}{K}$.
\begin{proof}
The proof is similar to Malik and Pitt~\shortcite{malik2011particle}. Note that $\pi^k$ is the normalized weight,
\begin{equation*}
    \pi_{t}^k=\frac{\omega_{t}^k}{\sum_{i=1}^{K}\omega_{t}^i}=\frac{p(y_t|\tilde{h}_t^k)}{\sum_{i=1}^K p(y_t|\tilde{h}_t^i)}
\end{equation*}
As $\frac{1}{K}{\sum_{i=1}^K p(y_t|\tilde{h}_t^i)} \rightarrow p(y_t|\theta)$,hence we have 
\begin{equation*}
\mathop{\max}_{k\in \{1,\dots,K\}}\lbrace \frac {\pi^k}{2} \rbrace \rightarrow \frac{\mathop{\max}_{k\in \{1,\dots,K\}}\lbrace p(y_t|\tilde{h}_t^k) \rbrace}{K p(y_t|\theta)}.
\end{equation*}
Because $p(y_t|\tilde{h}_t^k)$ is learned by the neural networks and has an upper bound so we have $\mathop{\max}_{k\in \{1,\dots,K\}}\lbrace p(y_t|\tilde{h}_t^k) \rbrace \leq U$, where $U$ is a large positive value.
\begin{equation*}
\lVert \widehat{F}_K(h)-\widetilde{F}_K(h) \rVert_\infty \rightarrow \frac{U}{K p(y_t|\theta)}.
\end{equation*}
 Hence, the distance is of order $\frac{1}{K}$.
\end{proof}
The difference between $\widetilde{F}(h)$ and $\widehat{F}_N(h)$ is displayed in Figure. \ref{fig:cont}. The computational overhead is in $O(N \times K \times N \times \log K)$ due to the necessary sorting of the sampled $\widetilde{h}_t$. 




We now summarize the steps of the particle filtering for LSTM model in period $t$ as shown in Algorithm~\ref{alg:algorithm1}. We would like to use a single non-linear function $f_1, f_2$ to represent our proposed LSTM with particle filtering layer described using Eqn~\ref{eq:lstm_1} to~\ref{eq:lstm_5} and Eqn~\ref{eq:pf_lstm}.
\begin{algorithm}[tb]
\caption{Particle Filtering for LSTM}
\label{alg:algorithm1}
\textbf{Input}: $K$ particles $\{ h_{t-1}^1,\dots,h_{t-1}^K \}$, driving exogenous information $\mathbf{x}_t$ and historical data $y_{t-1}$\\
\textbf{Parameter}: $\mathbf{W}_f$,$\mathbf{W}_i$,$\mathbf{W}_o$,$\mathbf{W}_C$,$\mathbf{b}_f$,$\mathbf{b}_i$,$\mathbf{b}_o$,$\mathbf{b}_C$ in the LSTM model, and $f_{noise}()$,$f_{weight}()$,$f_{out}()$ in the neural networks\\
\textbf{Output}: $N$ particles $\{ (h_{t}^1,c_{t}^1),\dots,(h_{t}^N,c_{t}^N) \}$ in the next timestep
\begin{algorithmic}[1] 
\STATE Generate the noise term $\epsilon_t$ in the transition update process;
\STATE Update $\{ h_{t-1}^k\}$ through the LSTM layer: \\$\widetilde{h}_t^k = f_1(h_{t-1}^k,\mathbf{x}_t)$;
\STATE Calculate the corresponding weights $\{ \pi_t^k \}$ for the hidden states;
\STATE Prediction the target value:
$\hat{y} = f_{out}(\bar{\mathbf{h}}_t)$;
\STATE Continuous resampling to get updated particles $\{h_t^k \}$;
\STATE \textbf{return} updated particles
\end{algorithmic}
\end{algorithm}

\begin{table*}[bt]
\centering
\begin{tabular}{cccc}
\toprule
Model                       & MAE & MAPE$\times10^{-1}\%$ & RMSE \\ \midrule
RNN      &$1.626\pm0.182$   & $0.140\pm0.0252$           &$3.995\pm0.175$         \\
CPF-RNN($K=10$)                     &$1.341\pm0.165$   & $0.101\pm0.0231$                   & $3.720\pm0.159$   \\
CPF-RNN($K=50$)                     & $1.184\pm0.041$   & $0.079\pm0.0054$                    & $3.574\pm0.044$    \\
DA-RNN                      & $1.349\pm0.117$   & $0.022\pm0.0164$                   &$2.183\pm 0.115$   \\
CPF-Encoder DA-RNN($K=10$)            &$1.153\pm0.043$  & $0.020\pm0.0014$                    & $1.876\pm0.078$    \\
CPF-Encoder DA-RNN($K=50$)           &$0.871\pm0.055$  & $0.015\pm0.0011$                   & $1.606\pm0.017$    \\
CPF-Decoder DA-RNN($K=10$)           & $0.813\pm0.026$   & $0.014\pm0.0006$                   & $1.550\pm0.024$   \\
CPF-Decoder DA-RNN($K=50$)           & \textbf{0.807 $\pm$ 0.012}   & \textbf{0.014 $\pm$ 0.0005}                 & \textbf{1.538 $\pm$ 0.008}  \\
CPF DA-RANN($K=10$)   & $0.811\pm0.030$   &  $0.014\pm0.0005$                     &$1.543\pm0.032$   \\
CPF DA-RANN($K=50$)   &  \textbf{0.808 $\pm$ 0.013}   & \textbf{0.014$ \pm$ 0.0005}                 & \textbf{1.529 $\pm$ 0.011}    \\\bottomrule
\end{tabular}
\caption{Time series prediction results over the NASDAQ 100 Stock Dataset.}
\label{tab:result}
\end{table*}

\subsection{Extensions to SOTA methods}
In this section, we will show how our proposed model can be incorporated into some SOTA methods, specifically the dual-stage attention-based RNN introduced by Qin {\it et al.}~\shortcite{qin2017dual}. The detailed model can be found in the appendix.
\subsubsection{Encoder with Input Attention} 
The encoder is applied to learn a mapping from the input sequence $\mathbf{x}_t$ to the hidden state $\mathbf{h}_t$ of the encoder at time $t$, where $\mathbf{x}_t\in\mathbb{R}^n$ and $\mathbf{h}_t\in \mathbb{R}^m$, $n$ is the number of driving (exogenous) series, and $m$ is the size of the hidden state. We use the same attention framework but replace the standard LSTM model with our proposed CPF-LSTM as the non-linear mapping function to capture the long-term dependencies of time series. More specifically, 
    \begin{equation}
        \mathbf{h}_t^k = f_1(\mathbf{h}_{t-1}^k, \tilde{\mathbf{x}}_t);
    \end{equation}
where $\tilde{\mathbf{x}}_t=(\alpha_1 x_t^1,\alpha_2 x_t^2,\dots,\alpha_n x_t^n)^\top$ is the encoded input to the Encoder. Again, we use the average value of the hidden state $\bar{\mathbf{h}}_{t-1} = \frac{1}{K}\sum_{k=1}^K \mathbf{\tilde{h}}_{t-1}^k$ to calculate the attention weight $\alpha_t^k$ of the $k$-th input driving series at time $t$. For the precise form of $\alpha$ and the encoder-decoder form, we refer the reader to Qin {\it et al.}~\shortcite{qin2017dual}.
\subsubsection{Decoder with Temporal Attention} 
The decoder is used to decode the encoded input information, which is the average value of the encoder hidden state $\bar{\mathbf{h}}_t$. Similarly, we keep the decoder mechanism the same but replace the LSTM model with CPF-LSTM, and use the mean of the decoder hidden state $\bar{\mathbf{d}}_{t-1} = \frac{1}{K}\sum_{k=1}^K \mathbf{\tilde{d}}_{t-1}^k$ to compute the attention weight and the context vector $\mathbf{c}_t$ at $t$. Then we use the weighted summed context vectors combined with the given target series $(y_1,y_2,\dots,y_{T-1})$ to compute the decoder input $\tilde{y}_{T-1}$. We update a set of particles to approximate the hidden state of the decoder:
\begin{equation}
    \mathbf{d}_t^k = f_2 (\mathbf{d}_{t-1}^k, \tilde{y}_{t-1})
\end{equation}
For the details, please refer to~\cite{qin2017dual}.

\subsubsection{Prediction}
Similar to Qin {\it et al.}~\shortcite{qin2017dual}, we use the combination of the decoder hidden state and the context vector to predict the target time series value, which is 
\begin{equation}
    \hat{y}_T = f_{out}(\mathbf{\bar{d}}_T, \mathbf{c}_T) 
\end{equation}

\subsubsection{Training Procedure}  
For purposes of easy computation, we only rely on the task-oriented objective function $\mathcal{O}(y_T,\hat{y}_T)$ to optimize the parameters, because the derivation of the ELBO on the likelihood is complex in the encoder-decoder framework. 

\section{Experiments}\label{sec:exp}
\subsection{Dataset}
We choose the NASDAQ 100 Stock dataset which is also used in~\cite{qin2017dual}. There are $81$ major corporation prices under NASDAQ 100, which are used as the exogenous information. The index value of the NASDAQ 100 is used as the target time series. The frequency of the data collection is minute-by-minute. Refer to~\cite{qin2017dual} for more details on the dataset. We use the first $80\%$ of the data points for training, and use the rest as the test dataset.

\subsection{Experiment Settings}
Our experimental settings are similar to the settings used in~\cite{qin2017dual}. The number of time steps in the window is $T=10$, and the size of hidden states for encoder $(m)$ and decoder $(p)$ are $m=p=128$. The number of particles is the same for CPF-LSTM in both the encoder and decoder. For all approaches, we train them 5 times and report their average performance and standard deviations to measure the effectiveness of various methods for time series prediction. We choose the mean absolute error (MAE), mean absolute percentage error (MAPE), and root mean square error (RMSE) as our metrics. They are defined as:
\begin{equation*}
    \begin{aligned}
    \text{MAE} &= \frac{1}{N}\sum_{i=1}^N |y_t^i -\hat{y}_t^i|\\
    \text{MAPE} & =\frac{1}{N}\sum_{i=1}^N \lvert \frac{y_t^i -\hat{y}_t^i}{y_t^i} \rvert \times 100\% \\
    \text{RMSE} &= \sqrt{\frac{1}{N}\sum_{i=1}^N (y_t^i -\hat{y}_t^i)^2}.
    \end{aligned}
\end{equation*}

\subsection{Results Comparison}
In table~\ref{tab:result}, we compare our model and its extensions to SOTA. The first row is the basic LSTM model. CPF-RNN is the LSTM model combined with the continuous particle filtering algorithm. DA-RNN is the dual-stage attention-based RNN introduced by~\cite{qin2017dual}, which also served as our baseline. CPF-Encoder and CPF-Decoder are the DA-RNN considering using our proposed CPF-LSTM model in encoder and decoder, respectively. CPF-DA-RNN considers the CPF-LSTM model in both the encoder and decoder. $K$ is the number of particles to approximate the hidden states in the particle filtering algorithm.

In particular, we find 
\begin{enumerate}
    \item When incorporated with the continuous particle filtering algorithm, it will outperform the original model. Specifically, SOTA combined with particle filtering in both the encoder and decoder has the best performance among all the models, and SOTA combined with particle filtering in the decoder has the lowest variance.
    \item Comparing the same model with the different numbers of particles, we observe that more particles improve the performance. Intuitively, more particles could lead to a better approximation of the complex belief distributions.
\end{enumerate}

\section{Conclusion}
In this paper, we propose a CPF-LSTM model and show how it can be extended to LSTM- based attention mechanism, which can achieve better performance.

\appendix
\section{Extensions}
In this section, we will show how our proposed model can be incorporated in some SOTA methods, i.e., the dual-stage attention-based RNN introduced by Qin {\it et al.}~\shortcite{qin2017dual}.
\subsubsection{Encoder with Input Attention} 
The encoder is applied to learn a mapping from the input sequence $\mathbf{x}_t$ to the hidden state $\mathbf{h}_t$ of the encoder at time $t$, where $\mathbf{x}_t\in\mathbb{R}^n$ and $\mathbf{h}_t\in \mathbb{R}^m$, $n$ is the number of driving (exogenous) series, and $m$ is the size of hidden state. We use the same attention framework but replace the standard LSTM model with our proposed CPF-LSTM as the non-linear mapping function to capture the long-term dependencies of time series. More specifically, 
    \begin{equation}
        \mathbf{h}_t^k = f_1(\mathbf{h}_{t-1}^k, \tilde{\mathbf{x}}_t);
    \end{equation}
where $\tilde{\mathbf{x}}_t=(\alpha_1 x_t^1,\alpha_2 x_t^2,\dots,\alpha_n x_t^n)^\top$ is the encoded input to the Encoder. Again, we use the average value of the hidden state $\bar{\mathbf{h}}_{t-1} = \frac{1}{K}\sum_{k=1}^K \mathbf{\tilde{h}}_{t-1}^k$ to calculate the attention weight $\alpha_t^k$ of the $k$-th input driving series at time $t$. For the precise form of $\alpha$ and the encoder-decoder form, we refer the reader to Qin {\it et al.}~\shortcite{qin2017dual}.
\subsubsection{Decoder with Temporal Attention} 
The decoder is used to decode the encoded input information, which is the average value of the encoder hidden state $\bar{\mathbf{h}}_t$. Similarly, we keep the decoder mechanism same but replace the LSTM model with CPF-LSTM, and use the mean of the decoder hidden state $\bar{\mathbf{d}}_{t-1} = \frac{1}{K}\sum_{k=1}^K \mathbf{\tilde{d}}_{t-1}^k$ to compute the attention weight and the context vector $\mathbf{c}_t$ at $t$. Then we use the weighted summed context vectors combined with the given target series $(y_1,y_2,\dots,y_{T-1})$ to compute the decoder input $\tilde{y}_{T-1}$. We update a set of particles to approximate the hidden state of the decoder:
\begin{equation}
    \mathbf{d}_t^k = f_2 (\mathbf{d}_{t-1}^k, \tilde{y}_{t-1})
\end{equation}
For the details, please refer to~\cite{qin2017dual}.

\subsubsection{Prediction}
Similar to~\cite{qin2017dual}, we use the combination of the decoder hidden state and the context vector to predict the target time series value, which is 
\begin{equation}
    \hat{y}_T = f_{out}(\mathbf{\bar{d}}_T, \mathbf{c}_T) 
\end{equation}

\subsubsection{Training Procedure}  
For purposes of easy computation, we only rely on the task-oriented objective function $\mathcal{O}(y_T,\hat{y}_T)$ to optimize the parameters, because the derivation of the ELBO on the likelihood is complex in the encoder-decoder framework. 






\bibliographystyle{named}
\bibliography{ijcai21}

\end{document}